\documentclass[conference]{IEEEtran}

\usepackage{graphicx}      
\usepackage{amsmath}
\usepackage{amsfonts}
\usepackage{algorithm}
\usepackage{algpseudocode}
\usepackage[hidelinks]{hyperref}

\IEEEoverridecommandlockouts 

\begin{document}
	
	\title{Visual Servoing Based on 3D Features: Design and Implementation for Robotic Insertion Tasks}
	
	\author{Antonio Rosales$^{1}$, Tapio Heikkilä$^{1}$, and Markku Suomalainen$^{1}$
        \thanks{This project has received funding from the ECSEL Joint Undertaking (JU) under Grant Agreement No 101007311. The JU receives support from the European Union’s Horizon 2020 research and innovation programme and Netherlands, Czech Republic, Spain, Greece, Ireland, Italy, Belgium, Latvia, Portugal, Germany, Finland, Romania and Switzerland}
		\thanks{$^{1}$A. Rosales, T. Heikkilä and M. Suomalainen are with the VTT Technical Research Centre of Finland Ltd, Oulu, Finland.
			{\tt\small antonio.rosales@vtt.fi}}
	}

	\maketitle
	
	\begin{abstract}
This paper proposes a feature-based Visual Servoing (VS) method for insertion task skills. A camera mounted on the robot's end-effector provides the pose relative to a cylinder (hole), allowing a contact-free and damage-free search of the hole and avoiding uncertainties emerging when the pose is computed via robot kinematics. Two points located on the hole's principal axis and three mutually orthogonal planes defining the flange's reference frame are associated with the pose of the hole and the flange, respectively. The proposed VS drives to zero the distance between the two points and the three planes aligning the robot's flange with the hole's direction. Compared with conventional VS where the Jacobian is difficult to compute in practice, the proposed featured-based uses a Jacobian easily calculated from the measured hole pose. Furthermore, the feature-based VS design considers the robot's maximum cartesian velocity. The VS method is implemented in an industrial robot and the experimental results support its usefulness.
	\end{abstract}
	

\section{Introduction}
Visual Servoing (VS) methods are a way to integrate eyesight on robots since the main element of these methods is a camera providing visual information. The camera can be mounted on the robot's end-effector to have an eye-in-hand approach or fixed in the world observing the robot's end-effector and the target object to have an eye-to-hand approach \cite{b:CorkeVCbook97}.

Considering the type of visual features used in VS, there are two classes of VS, Image-Based Visual Servoing (IBVS) and Position-Based Visual Servoing (PBVS). IBVS compares purely 2D image features, and PBVS compares poses, i.e., 3D information of the object. The VS method presented in this paper is PBVS-type since it acts on the robot's pose relative to the object using data from a 3D camera to avoid the main drawbacks of IBVS such as lack of depth information and illumination dependency. Moreover, 3D camera's data has low sensitivity to illumination changes, and variation in the object's position (rotation and translation) \cite{b:Prasad2006FirstSI}.

Insertion tasks are often required in robotic applications with uncontrolled environments such as construction, mining, and services. For example, service robots filling gasoline into car's tanks, robots charging booms in mining applications, or robots filling holes with concrete, see \cite{b:bi2020automatic,b:chen2022computer,b:BonchisTASE14}. Note that there should be a hole identification stage in the mentioned insertion task. There are three main approaches to identifying the hole and its direction.  One approach uses force sensors to find the holes through contact, another uses a camera to find the pose of the hole, and the third is a combination of both. The identification step is mainly done through vision when having contact with the environment is not desired. In this paper, the insertion task is executed using purely visual feedback to avoid damage to the hole/object/flange when the hole is searched through contact.

Plenty of research studies have proposed visual servoing solutions for insertion tasks. The diversity of methods goes from classical object detection and control methods presented in the seventies, see \cite{b:SHIRAI197399}, to recently presented advanced methods based in Deep Learning, see \cite{b:HaugaarDPML21} and \cite{b:TriyonoputroIROS19}. However, the velocity limits of the robot end-effector are barely considered when the visual servoing is designed. In PBVS, the most frequent recommendation is to choose the gain bigger than zero \cite{b:Chaumette06}, but how to set this gain to consider the task space velocity limits is barely addressed.

Distance minimization between 3D features like planes and points is commonly used in robot and hand-eye calibration, and target localization. In \cite{b:heikkila2010interactive}, our research team used minimization of the distances between measured 3D points and reference surfaces to determine the target object pose. A hand-eye calibration method minimizing the distance between points is presented in \cite{b:XingTII23}. In \cite{b:PetersJFR23}, a robot calibration method based on point-to-plane distance minimization is presented. Despite the mentioned methods end-up computating translational and rotational corrections of the robot's end-effector, these featured-based methods are rarely used for visual servoing applications. A visual servoing using two-dimensional features is presented in \cite{b:MorrowICRA97}, however the 3D features case is not addressed.

This paper presents a visual servoing based on 3D features such as points and planes. The task of locating the robot's flange at a desired pose with respect to a hole's reference frame exemplifies the feature-based VS design. A 3D camera provides the data to estimate the flange's pose relative to the hole's reference frame. Two points and three planes are associated with the hole direction and the flange pose, respectively. Then, the proposed feature-based VS generates the robot's corrections required to drive to zero the distance between the two points and the three plains, aligning the robot's flange with the hole. The Jacobian used to calculate the robot's corrections is computed directly from pose measurements, then, the proposed method does not present the difficulties of classical VS to compute the Jacobian in practice \cite{b:HutchinstonCorke96,b:PrzystupaICRA21}. Furthermore, the featured-based VS design provides robot's corrections that fit the robot task space velocity limits. Experimental results showing the utility of the proposed VS method are presented.

The structure of the paper is the following. Section II presents the details of the VS based on distances between points and planes. Section III presents the description of the task and the details of the VS implementation.  The experimental results, discussion, and conclusions are shown in Sections IV, V, and VI, respectively.


\section{Feature-based Visual servoing}\label{sec:VS2}
The main goal of this VS method is to align a peg center line to the center line of a hole at a desired level from the top plane of the hole surroundings. The representation and minimization of distances between 3D lines, and even between 3D points and lines, results in non-linear models. A linear model and better correspondence can be attained by representing one of the 3D lines with two perpendicular planes. Then, one center line is a 3D line defined by the intersection of two planes, and the other is represented by two 3D points forming a 3D line \cite{b:HeikkilaSICE88}. Then, when aligning the flange with the hole, we end up minimizing distances between 3D points and 3D planes. 

\begin{figure}[h] 
	\centering
	\includegraphics[width=9cm]{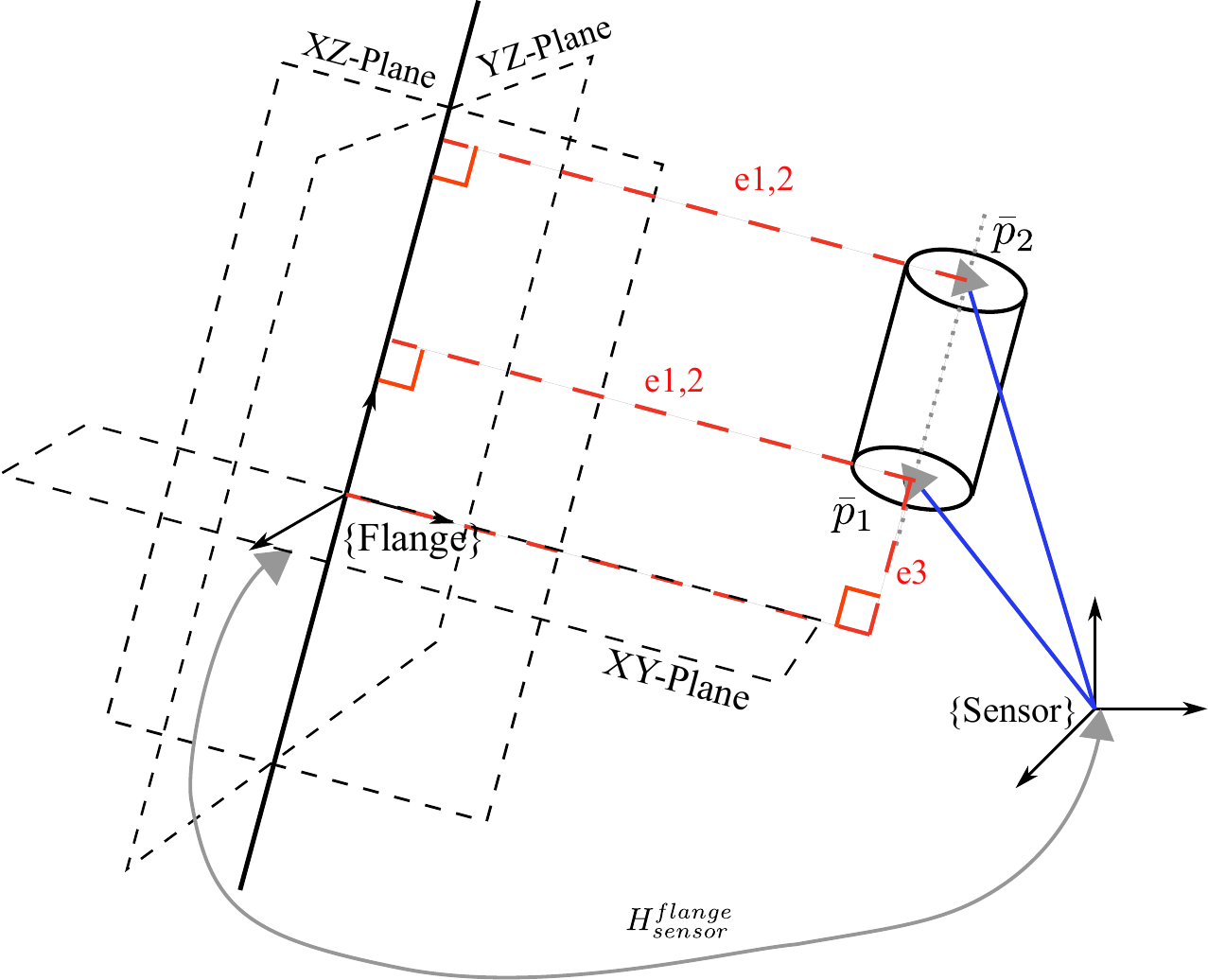}
	\caption{Distance from two points to three planes}
	\label{fig:Planes2Points}
\end{figure}

The three planes are choosen to be on the YZ-plane, XZ-plane, and XY-plane of the hose/flange's reference frame since planes YZ and XZ form the flange center line and the plane XY is used to define the distance between the flange and the hole, see Fig. \ref{fig:Planes2Points}.

The two points are put on the hole center line. One point represents the hole's location, and the second point defines the hole's direction, see $\bar{p}_1$ and $\bar{p}_2$ in Fig. \ref{fig:Planes2Points}.

The error to be minimized is composed of the distances from a point $p_i$ to a plane $N_j$,
\begin{equation*}
	e_{i,j}=\bar{n}_j\cdot\bar{p}_i
\end{equation*}
where $\bar{n}_j$ is an unitary vector normal to the plain $N_j$, and $\bar{p}_i=(p_{i_x},p_{i_y}, p_{i_z})$ is the point of interest. In this case, there are 3 planes and two points, then the indexes are $i=1,2$, and $j=1,2,3$. Here and along the paper, we are considering $N_1$, $N_2$, and $N_3$ are YZ-plane, XZ-plane, and XY-plane, respectively. The point $\bar{p}_1$ represents the hole's location and $\bar{p}_2$ is defined by the sum of $p_1$ plus a unitary vector ($\bar{n}_{h_{dir}}$) in the direction of the hole's center line i.e. $\bar{p}_2=\bar{p}_1+c\bar{n}_{h_{dir}}$, where $c>0$. For example, the error of $\bar{p}_1$ with respect to plane $N_1$ is $e_{1,1}=\bar{n}_1\cdot\bar{p}_1$.

Considering the vector of parameters  $\bar{x}=[x \ y \ z \ b \ c]$, where $(x,y,z)$ and $(b,c)$ are the translations and Euler angles of the flange/hose, respectively. Note that the vector $\bar{x}$ does not include the angle $a$ associated with rotation around z-axis since the rotation around $z$ does not provide information on the alignment of the flange with the hole.

The increments of the flange/hose $\Delta\bar{x}=[\Delta x \ \Delta y \ \Delta z \ \Delta b \ \Delta c]$ can be computed via the next equation \cite{b:HeikkilaSPIE99},
\begin{equation}\label{eq:DeltaX}
	\Delta\bar{x}=J^{-1}\bar{e}
\end{equation}
where $\bar{e}=[e_{1,1} \ e_{1,2} \ e_{2,1} \ e_{2,2} \ e_{1,3}]$ is a column vector composed with the combinations of $e_{i,j}$, and the Jacobian $J$ is defined as,
\begin{equation}
	J=\frac{\partial e_{i,j}}{\partial \bar{x}}=\frac{\partial e_{i,j}}{\partial p_i}\frac{\partial p_i}{\partial \bar{x}}.
\end{equation}
For the case $e_{1,1}$, the partial derivative is 
\begin{eqnarray*}
	\frac{\partial e_{1,1}}{\partial \bar{x}}&=&
		\frac{\partial e_{1,1}}{\partial p_1}\frac{\partial p_1}{\partial \bar{x}}=\begin{bmatrix}
	1 \ 0 \ 0
\end{bmatrix}\begin{bmatrix}
\frac{\partial p_{1x}}{\partial \bar{x}} \\
\frac{\partial p_{1y}}{\partial \bar{x}} \\
\frac{\partial p_{1z}}{\partial \bar{x}}
\end{bmatrix}\\
&=&\begin{bmatrix}
1 \ 0 \ 0
\end{bmatrix}\begin{bmatrix}
\frac{d p_{1x}}{dx} & \frac{d p_{1x}}{dy} & \frac{d p_{1x}}{dz} & \frac{d p_{1x}}{db} & \frac{d p_{1x}}{dc} \\
\frac{d p_{1y}}{dx} & \frac{d p_{1y}}{dy} & \frac{d p_{1y}}{dz} & \frac{d p_{1y}}{db} & \frac{d p_{1y}}{dc} \\
\frac{d p_{1z}}{dx} & \frac{d p_{1z}}{dy} & \frac{d p_{1z}}{dz} & \frac{d p_{1z}}{db} & \frac{d p_{1x}}{dc}
\end{bmatrix},\\
&=&\begin{bmatrix}
1 \ 0 \ 0
\end{bmatrix}\begin{bmatrix}
1 & 0 & 0 & p_{1_z} & -p_{1_y} \\
0 & 1 & 0 & 0 & p_{1_x} \\
0 & 0 & 1 & -p_{1_x} & 0
\end{bmatrix}
\end{eqnarray*}
and for the other errors,
\begin{eqnarray*}
	\frac{\partial e_{1,2}}{\partial \bar{x}}&=&\begin{bmatrix}
0 \ 1 \ 0
\end{bmatrix}\begin{bmatrix}
1 & 0 & 0 & p_{1_z} & -p_{1_y} \\
0 & 1 & 0 & 0 & p_{1_x} \\
0 & 0 & 1 & -p_{1_x} & 0
\end{bmatrix}\\
    \frac{\partial e_{2,1}}{\partial \bar{x}}&=&\begin{bmatrix}
1 \ 0 \ 0
\end{bmatrix}\begin{bmatrix}
1 & 0 & 0 & p_{2_z} & -p_{2_y} \\
0 & 1 & 0 & 0 & p_{2_x} \\
0 & 0 & 1 & -p_{2_x} & 0
\end{bmatrix}\\
    \frac{\partial e_{2,2}}{\partial \bar{x}}&=&\begin{bmatrix}
0 \ 1 \ 0
\end{bmatrix}\begin{bmatrix}
1 & 0 & 0 & p_{2_z} & -p_{2_y} \\
0 & 1 & 0 & 0 & p_{2_x} \\
0 & 0 & 1 & -p_{2_x} & 0
\end{bmatrix}\\
    \frac{\partial e_{1,3}}{\partial \bar{x}}&=&\begin{bmatrix}
0 \ 0 \ 1
\end{bmatrix}\begin{bmatrix}
1 & 0 & 0 & p_{1_z} & -p_{1_y} \\
0 & 1 & 0 & 0 & p_{1_x} \\
0 & 0 & 1 & -p_{1_x} & 0
\end{bmatrix}
\end{eqnarray*}

Then, the total Jacobian $J$ is
\begin{equation*}
	J=\begin{bmatrix}
		\frac{\partial e_{1,1}}{\partial \bar{x}} \\
		\frac{\partial e_{1,2}}{\partial \bar{x}} \\
		\frac{\partial e_{2,1}}{\partial \bar{x}} \\
		\frac{\partial e_{2,2}}{\partial \bar{x}} \\
		\frac{\partial e_{1,3}}{\partial \bar{x}} 
	\end{bmatrix}=\begin{bmatrix}
	\frac{d p_{1x}}{dx} & \frac{d p_{1x}}{dy} & \frac{d p_{1x}}{dz} & \frac{d p_{1x}}{db} & \frac{d p_{1x}}{dc} \\
	\frac{d p_{2x}}{dx} & \frac{d p_{2x}}{dy} & \frac{d p_{2x}}{dz} & \frac{d p_{2x}}{db} & \frac{d p_{2x}}{dc} \\
	\frac{d p_{1y}}{dx} & \frac{d p_{1y}}{dy} & \frac{d p_{1y}}{dz} & \frac{d p_{1y}}{db} & \frac{d p_{1y}}{dc} \\
	\frac{d p_{2y}}{dx} & \frac{d p_{2y}}{dy} & \frac{d p_{2y}}{dz} & \frac{d p_{2y}}{db} & \frac{d p_{2y}}{dc} \\
	\frac{d p_{1z}}{dx} & \frac{d p_{1z}}{dy} & \frac{d p_{1z}}{dz} & \frac{d p_{1z}}{db} & \frac{d p_{1x}}{dc}
\end{bmatrix}
\end{equation*}

The corrections defined in equation (\ref{eq:DeltaX}) can be scaled using a constant $\beta$. Then, equation (\ref{eq:DeltaX}) can be rewritten as
\begin{equation}\label{eq:DeltaXv2}
	\Delta\bar{x}=\beta J^{-1}\bar{e}
\end{equation}
where $\beta>0$ has to be adequately selected to match the size of the corrections with the robot's velocities.

\subsection{Design of $\beta$}
Define $v_{max}$ and $w_{max}$ as the maximum translational and rotational velocity of the robot's flange.

The maximum translational increment can be computed from $v_{max}$ as,
\begin{equation}
    \Delta \bar{x}_{t_{max}}=v_{max}\tau
\end{equation}
where $\tau$ is the sampling time, and the total translational increment can be obtained from,
\begin{equation}
    \Delta \bar{x}_{t_T}=\sqrt{\Delta x^2+\Delta y^2+\Delta z^2},
\end{equation}

A constant $\beta_{p}$ associated with the vector of translational corrections $[\Delta x \ \Delta y \ \Delta z]$ can be computed as follows
\begin{equation}
    \left\{
    \begin{matrix}
        \beta_{p}=\beta_{p}^{'}\Delta \bar{x}_{t_T} \ \text{if} \ \beta_{p}<\Delta \bar{x}_{t_{max}}\\
        \beta_{p}=\Delta \bar{x}_{t_{max}} \ \text{if} \ \beta_{p}\geq\Delta \bar{x}_{t_{max}}
    \end{matrix}\right.
\end{equation}
where $\beta_{p}^{'}>0$. Then, the corrections satisfying the translational velocity limits can be computed as follows
\begin{equation}\label{eq:velcorr}
    \begin{bmatrix}
        \Delta x_{robot} \\
        \Delta y_ {robot}\\
        \Delta z_{robot} \\
\end{bmatrix}=\begin{bmatrix}
        \frac{\beta_{p}}{\Delta \bar{x}_{t_T}}\Delta x \\
        \frac{\beta_{p}}{\Delta \bar{x}_{t_T}}\Delta y \\
        \frac{\beta_{p}}{\Delta \bar{x}_{t_T}}\Delta z \\
\end{bmatrix}
\end{equation}

For the rotational velocity limits, the maximum rotational increment is
\begin{equation}
    \Delta \bar{x}_{w_{max}}=w_{max}\tau
\end{equation}
where $\tau$ is the sampling time, and assuming small rotation increments the total rotational increment is
\begin{equation}
    \Delta \bar{x}_{r_T}=\sqrt{\Delta b^2+\Delta c^2},
\end{equation}

Using a constant $\beta_{r}$ for rotational corrections $[\Delta b \ \Delta c]$, $\beta_{r}$ is computed as follows
\begin{equation}
    \left\{
    \begin{matrix}
        \beta_{r}=\beta_{r}^{'}\Delta \bar{x}_{r_T} \ \text{if} \ \beta_{r}<\Delta \bar{x}_{r_{max}}\\
        \beta_{r}=\Delta \bar{x}_{r_{max}} \ \text{if} \ \beta_{r}\geq\Delta \bar{x}_{r_{max}}
    \end{matrix}\right.
\end{equation}
where $\beta_{r}^{'}>0$. The corrections satisfying the rotational velocity limits can be computed as follows
\begin{equation}\label{eq:rotcorr}
    \begin{bmatrix}
        \Delta b_{robot} \\
        \Delta c_ {robot}\\
\end{bmatrix}=\begin{bmatrix}
        \frac{\beta_{r}}{\Delta \bar{x}_{r_T}}\Delta b \\
        \frac{\beta_{r}}{\Delta \bar{x}_{r_T}}\Delta c \\
\end{bmatrix}
\end{equation}

Then, the corrections to be sent to the robot satisfying the provided velocity limits are $[\Delta x_{robot} \ \Delta y_{robot} \ \Delta z_{robot} \ \Delta b_{robot} \ \Delta c_{robot}]$.

\section{Features-based VS implementation}
\begin{figure}[h]
	\centering
	\includegraphics[width=8cm]{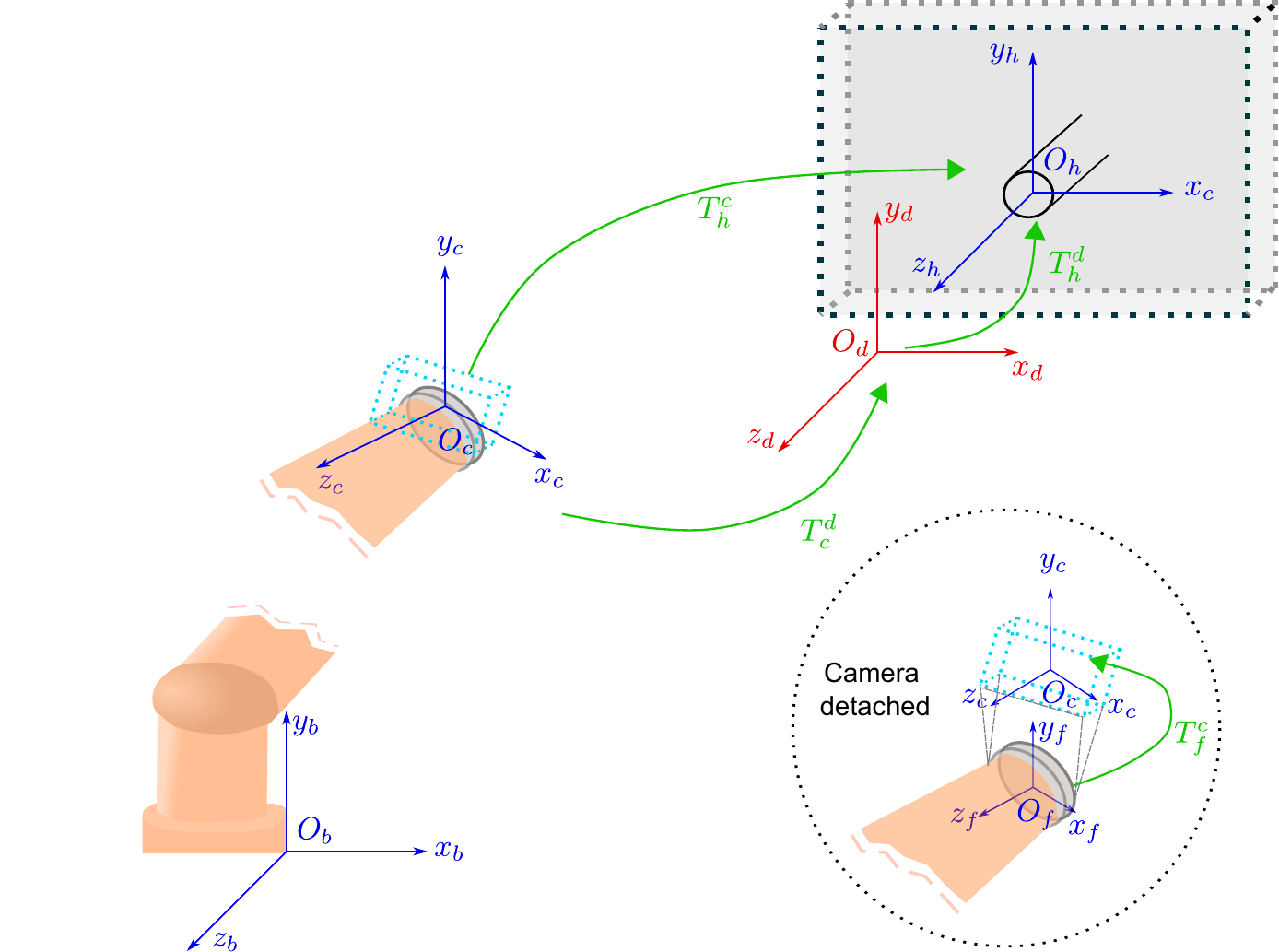}
	\caption{Description of the task, and reference frames for an eye-in-hand camera approach}
	\label{fig:RobVSframes}
\end{figure}

Consider a robot arm with a 3D camera mounted on its end-effector. A hole (object) is located inside the robot's task-space and the visible range of the camera, see Fig. \ref{fig:RobVSframes}.  We are assuming the camera is calibrated. The main task is to move the robot's end-effector close to the hole and to align the flange with the hole's direction i.e. to drive the camera/flange reference frame $O_c/O_f$ to the desired reference frame $O_d$. 

For implementing the feature-based VS, the first step is to identify/localize the object(hole), and the second step is to design the control input that drives the robot's end-effector to the desired pose. The next subsection contains the details of the implementation and the object localization method.

\subsection{Features-based VS implementation}
\begin{figure}[h] 
	\centering
	\includegraphics[width=9cm]{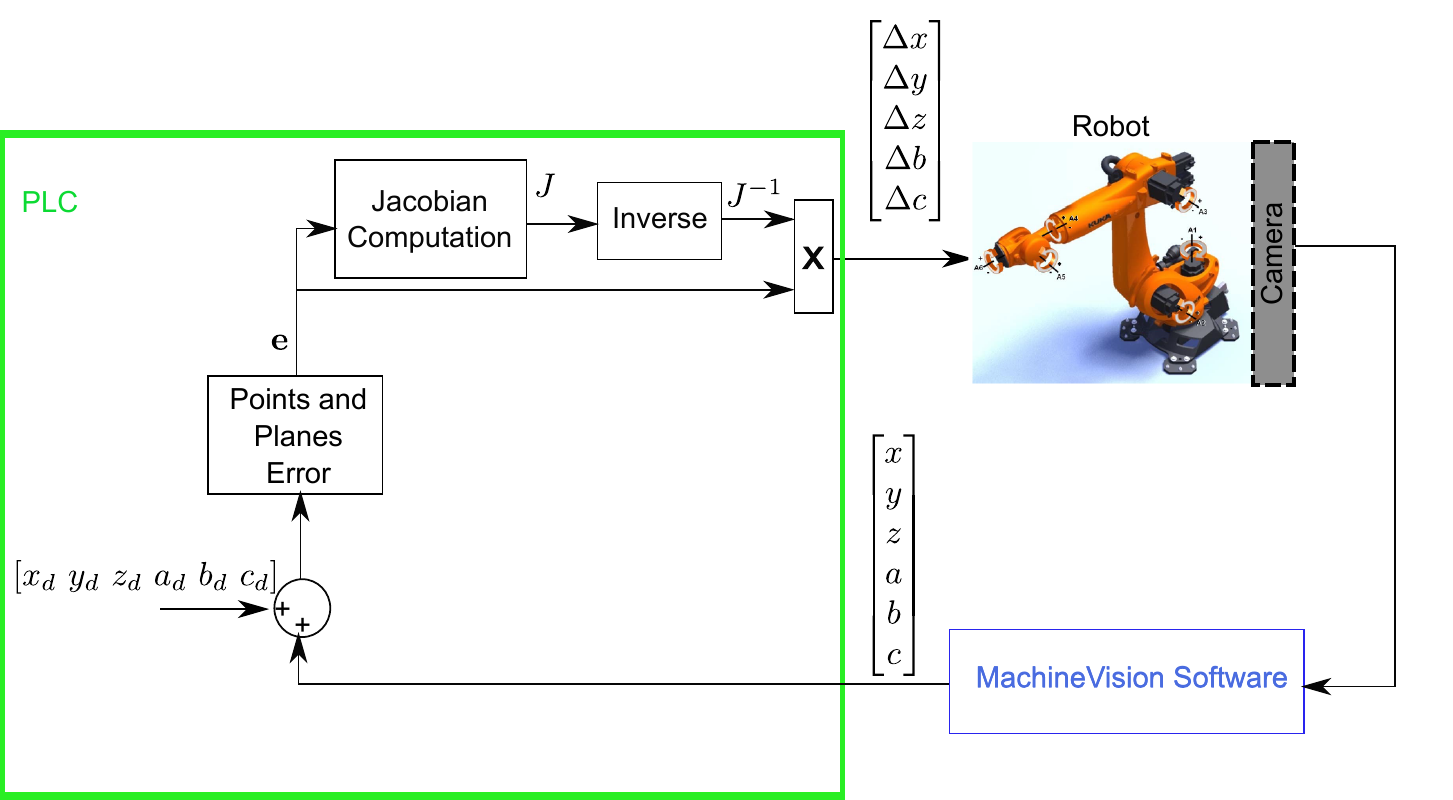}
	\caption{Block diagram of the Feature-based VS}
	\label{fig:bdImpl}
\end{figure}

Fig. \ref{fig:bdImpl} presents the block diagram of the proposed featured-based VS implementation. The model of the robot is KUKA KR120 R2500, and the camera is an Intel® RealSense D435 depth camera with a range from 0.3 to 3 meters. 

The block named MachineVision Software processes the point-cloud from the camera, and provides the position and orientation of camera's reference frame ($O_c$) relative to the hole's reference frame ($O_h$). This block contains a localization method developed by our research group, see \cite{b:Ahola2017} for more details.  

The block PLC is a programmable logic controller model Beckhoff’s CX5020. The PLC receives the position and orientation of the camera, and the feature's error $\mathbb{e}$ is computed by comparing the measured camera pose with the desired camera/flange pose. Then, the error is used to compute the control input,  i.e. translation and rotational corrections $(\Delta x,\Delta y,\Delta z,\Delta b,\Delta c)$,  to be sent to the robot. Note that the desired camera/flange pose is defined by the position and orientation values that align the camera/flange with the hole.

\subsection{Object localization and scanning}\label{sub:ObjReco}

Before running the VS for the first time, the area around the hole is scanned to find the best camera angles and distances for hole's identification. Once, these best angles and distances are found, they can be used to run the VS for other holes. 

\begin{figure}[h]
	\centering
	\includegraphics[width=7cm]{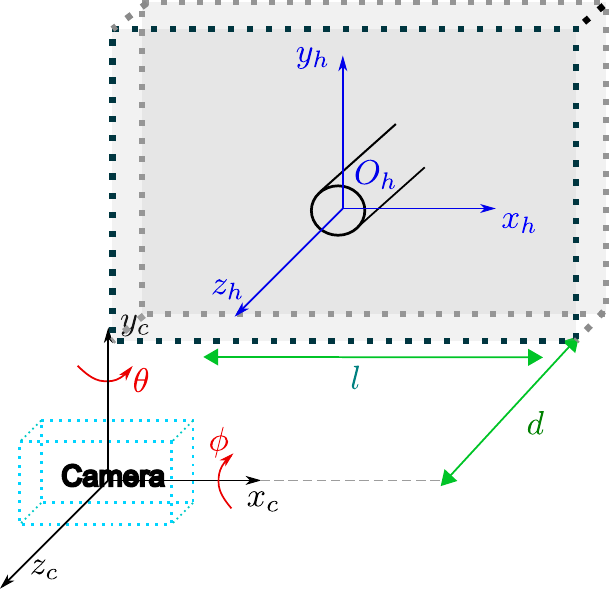}
	\caption{Parameters used to scan the area around the hole}
	\label{fig:Mapping2}
\end{figure}

Fig. \ref{fig:Mapping2} shows the parameters involved during the scanning. The parameter $d$ represents the distance from the camera to the hole along axis $z_h$, the parameter $l$ defines the movement along axis $x_h$, and the angles $\theta$ and $\phi$ represent the rotation of the camera along its $y$-axis and $x$-axis, respectively.

The scanning consists on variations of the distances $d$ and $l$, and angles $\theta$ and $\phi$. For example, assume the camera is located at the distances $d_0$ and $l_0$,  and three angle values will be tested i.e. $\theta_{0,1,2}$ and $\phi_{0,1,2}$. Then, there will be nine different view points at $(d_0,l_0)$ coming from the combinations $(\theta_{0},\phi_{0})$, $(\theta_{0},\phi_{1})$, $(\theta_{0},\phi_{2})$, $(\theta_{1},\phi_{0})$, $(\theta_{1},\phi_{1})$, $(\theta_{1},\phi_{2})$, $(\theta_{2},\phi_{0})$, $(\theta_{2},\phi_{1})$, and $(\theta_{2},\phi_{2})$. The next set of viewpoints will come when distance $d_0$(or $l_0$) changes to a new one $d_1$(or $l_1$). Therefore, a set of viewpoints will be tested at every camera location $(d_j,l_i)$, where $i$ and $j$ define the number of distances to be considered. If more data is required, the number of distances and angle points can be increased.

\subsubsection{Example of scanning}
Using the robot and the camera presented in Fig. \ref{fig:bdImpl}, The scanning is executed with the next parameters:
\begin{itemize}
	\item $d_{min}=30$~[cm], $d_{max}=120$~[cm], $d_{step}=30$~[cm]
	\item $x_{min}=0$~[cm], $x_{max}=45$~[cm], $x_{step}=15$~[cm]
	\item $\phi_{min}=-10^{\circ}$, $\phi_{max}=10^{\circ}$, $\phi_{step}=10^{\circ}$
	\item $\theta_{min}=-10^{\circ}$, $\theta_{max}=10^{\circ}$, $\theta_{step}=10^{\circ}$
\end{itemize}

\begin{figure}[h]
	\centering
	\includegraphics[width=9cm]{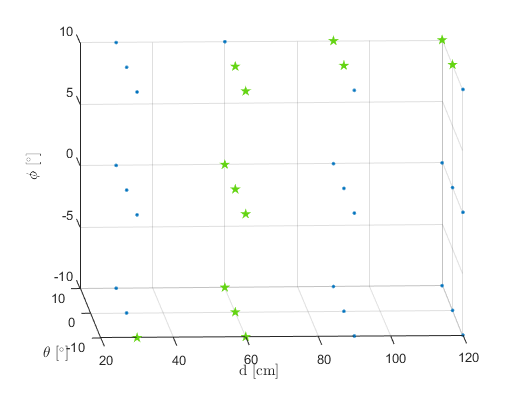}
	\caption{Data during scanning test: Green points indicate the hole was found}
	\label{fig:m1}
\end{figure}

Fig. \ref{fig:m1} presents the camera viewpoints during the test. Every point corresponds to a different viewpoint defined by the distances $d$ and $l$, and the combination of angles $\theta$ and $\phi$. The points marked with a star symbol ($\star$) indicate the viewpoints where the hole was found. From Fig. \ref{fig:m1}, one can conclude that the appropriate camera views to identify the cylinder are the ones with the parameters inside the next ranges: $30<d<120$ [cm], $-10<\theta<10$ [$^{\circ}$], and $-10<\phi<10$ [$^{\circ}$]

\section{Experimental results}
\begin{figure}[h] 
	\centering
	\includegraphics[width=9cm]{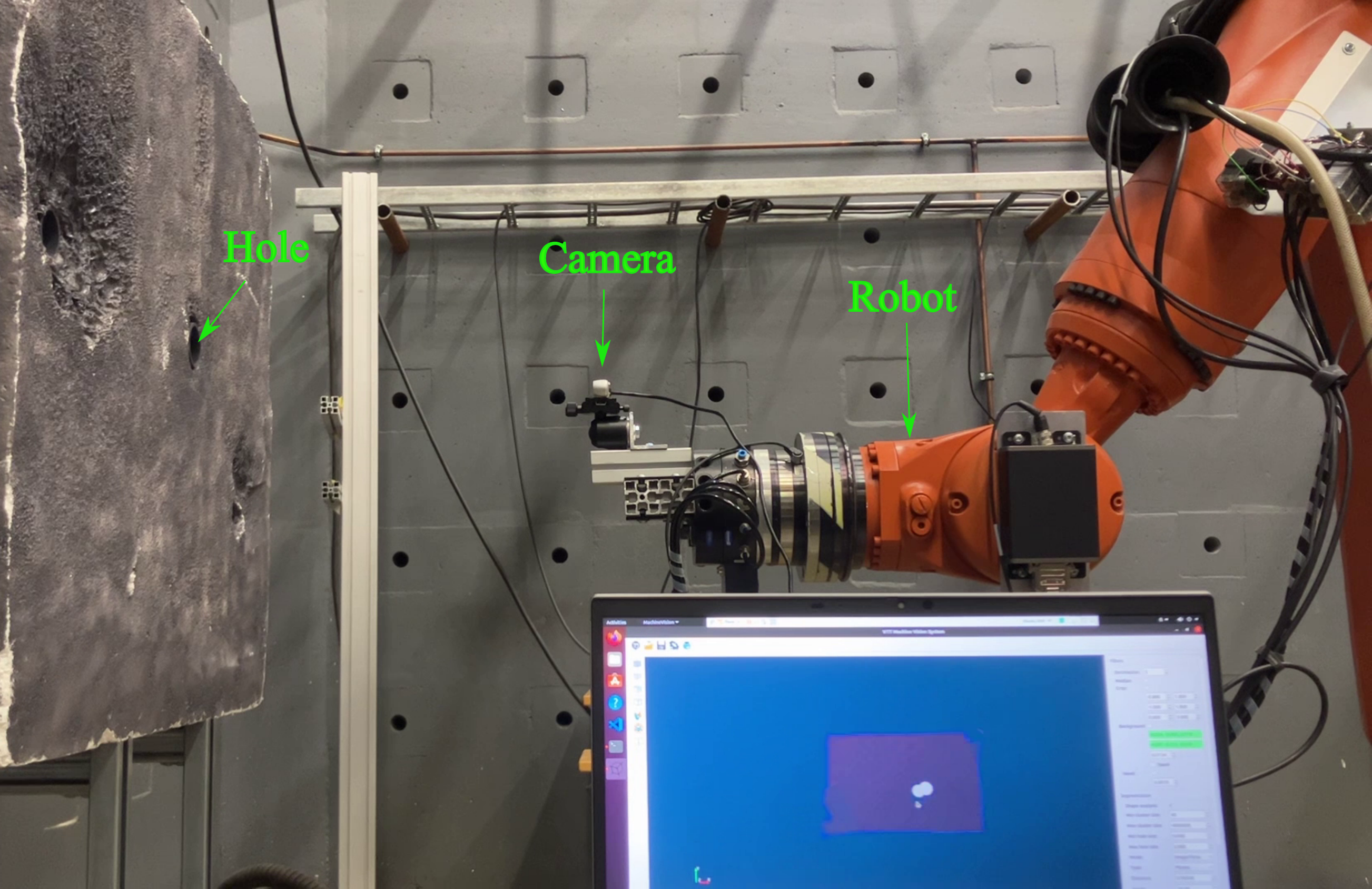}
	\caption{Experimental setup}
	\label{fig:RobotLab}
\end{figure}

Fig. \ref{fig:RobotLab} presents the experimental setup i.e. the robot, the camera, and the hole. The camera used for the experiments is the Intel RealSense D435 camera. The industrial robot is a KUKA KR-120, and the object is a hole drilled in a styrofoam wall.

The featured-based VS approach is implemented following the block diagram of Fig. \ref{fig:bdImpl} with a sampling time of $\tau=4$~miliseconds. Considering a $v_{max}=50$~[mm/s] and $w_{max}=40$~[degrees/s], the values maximum corrections are $\Delta \bar{x}_{p_{max}}=0.2$[mm], and $\Delta \bar{x}_{w_{max}}=0.16$[degrees]. The selected visual servoing gains are $\beta_{p}^{'}=0.001$, and $\beta_{r}^{'}=0.001$.

The feature-based VS is tested considering the initial camera pose $x=0.11$, $y=0.005$, $z=0.9$ [m], and $b=8$, $c=27$ degrees relative to the hole frame. The desired pose for the flange is $x_d=0$, $y_d=0.15$, $z_d=0.6$ [m], and $b=0$, $c=0$ degrees.\footnote{A video of the test can be found in the next link \href{https://drive.google.com/file/d/1l4IsQy1PeBBcyMTGd3UaQJXHeFQ_akZI/view?usp=sharing}{Video-Test}}

\begin{figure}[h] 
	\centering
	\includegraphics[width=9cm]{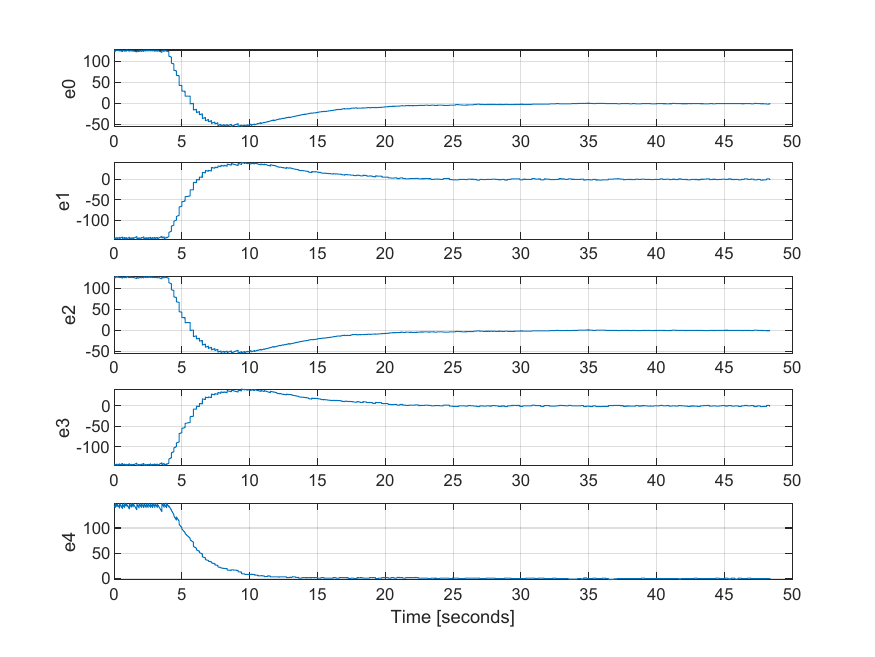}
	\caption{Distance errors of the featured-based VS}
	\label{fig:ExpFetErrors}
\end{figure}

Fig. \ref{fig:ExpFetErrors} presents the distance errors $\bar{e}$ of equation (\ref{eq:DeltaXv2}). Note that the VS starts after four seconds. During the first four seconds, the robot is at the initial conditions. Fig. \ref{fig:ExpFetErrors} shows that distance error $\bar{e}$ is driven to zero in 20 seconds.

\begin{figure}[h] 
	\centering
	\includegraphics[width=9cm]{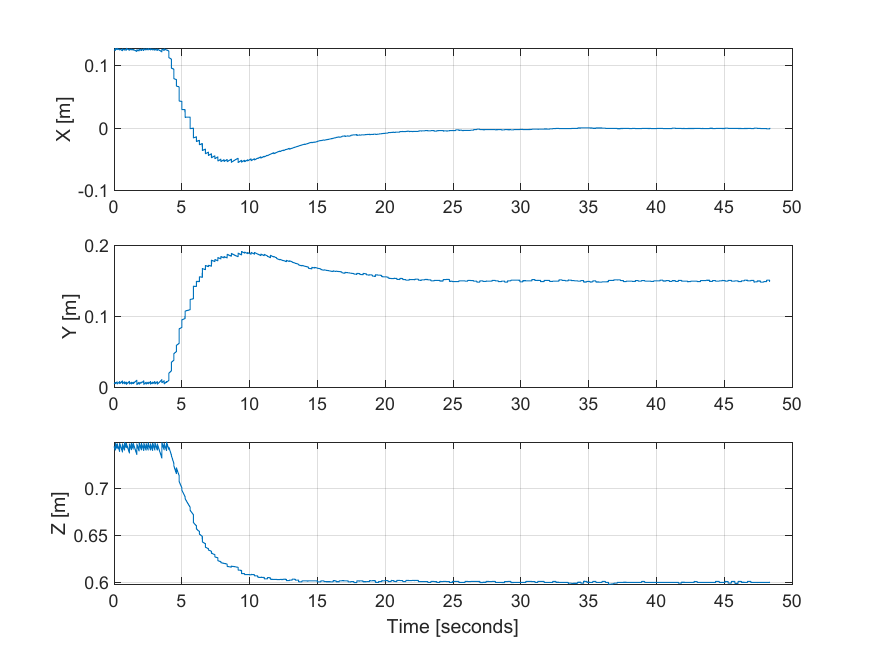}
	\caption{Tranlations XYZ }
	\label{fig:ExpFetXYZ}
\end{figure}

\begin{figure}[h] 
	\centering
	\includegraphics[width=9cm]{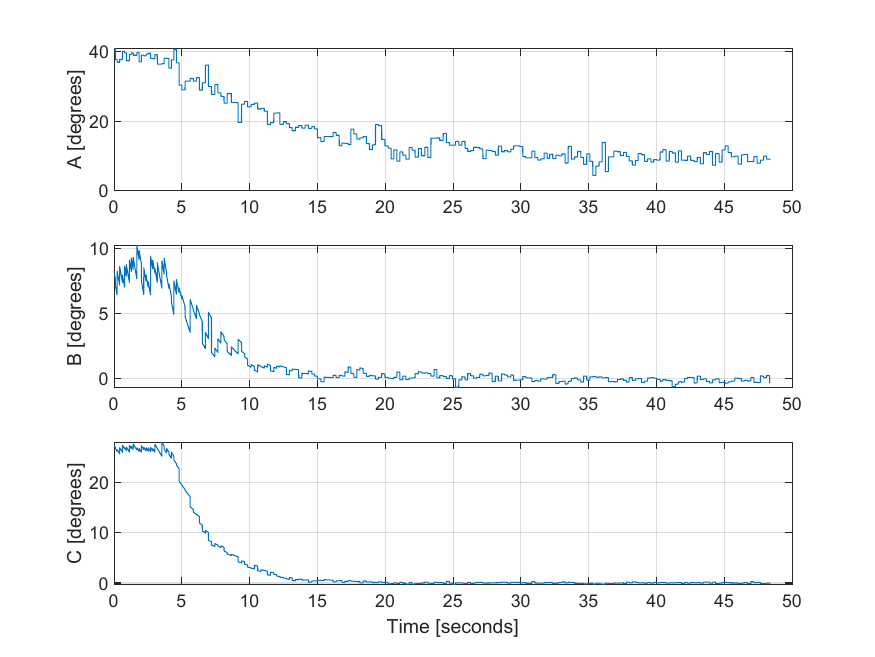}
	\caption{Rotation angles $a$,$b$,$c$: rotation $a$ is only illustration, no use in the control }
	\label{fig:ExpFetABC}
\end{figure}

Fig. \ref{fig:ExpFetXYZ} and \ref{fig:ExpFetABC} present the translations $x,y,z$ and rotations angles $b,c$ of the camera during the test. From Fig. \ref{fig:ExpFetXYZ} and \ref{fig:ExpFetABC}, it can be seen that the desired position $x_d=0$, $y_d=0.15$, $z_d=0.6$ [m], and the desired rotational angles $b=0$, $c=0$ are reached around 20 seconds. The rotation angle $a$ is shown in Fig. \ref{fig:ExpFetABC} but no correction is generated to this angle since it does not provide information about the orientation of the hole. From Fig. \ref{fig:ExpFetXYZ}, note that the translation on $z$ is approximately linear from four to ten seconds. Namely, the displacement on $z$ is 0.3~[m] (from 0.9 to 0.6) in six (from 4 to 10) seconds approximately. which is equivalent to 0.05~[m/second] and it match with the maximum velocity of $v_{max}=50$~[mm/s].

\section{Discussion}
To the knowledge of the authors, a visual servoing based on minimization of distances between 3D features has not been presented before. The visual servoing presented in \cite{b:MorrowICRA97} used 2D features but the 3D case is not considered. 3D features have been used for robot skills definition, robot calibration, and target localization, see \cite{b:MorrowICRA97,b:heikkila2010interactive,b:XingTII23,b:PetersJFR23} but not in visual servoing. The proposed feature-based VS uses a Jacobian computed easily from measured poses, it does not present the difficulties of classical VS to calculate the Jacobian in practice. Limitations of the featured-based VS might appear when the target object has complex geometry, making fitting 3D features difficult. However, the target object can be analyzed using points, edges, and surfaces in several industrial tasks, see \cite{b:MorrowICRA97}.

Considering the implementation of the featured-based VS, the scanning method presented in subsection \ref{sub:ObjReco} is useful for determining the best camera pose to identify the hole. The presented scanning procedure is a preliminary version. In future work, we plan to execute the scanning defining the percentage of identification success at every viewpoint.

The proposed featured-based visual servoing generates corrections satisfying the maximum correction that preserves the task-space velocity limits. Note that the velocity limits are considered indirectly via translational and rotational corrections. Future research is planned to add velocity and acceleration limits directly. 

\section{Conclusions}
This paper presents a feature-based visual servoing (VS) that minimizes the distance between two points and three plains. The presented VS is useful for applications that require alignment of the robot's flange with a hole. Furthermore, the design of the features-based VS ensures that the corrections to be sent to the robot do not surpass the robot's task-space velocity limits. The effectiveness of the proposed VS is tested via real-world experiments. The task executed during the experiments is to drive the robot's flange to a desired pose with respect to a hole, however, another task can be defined and executed using the same approach.

\bibliographystyle{IEEEtran}
\bibliography{bio}

\begin{thebibliography}{10}
\providecommand{\url}[1]{#1}
\csname url@samestyle\endcsname
\providecommand{\newblock}{\relax}
\providecommand{\bibinfo}[2]{#2}
\providecommand{\BIBentrySTDinterwordspacing}{\spaceskip=0pt\relax}
\providecommand{\BIBentryALTinterwordstretchfactor}{4}
\providecommand{\BIBentryALTinterwordspacing}{\spaceskip=\fontdimen2\font plus
\BIBentryALTinterwordstretchfactor\fontdimen3\font minus
  \fontdimen4\font\relax}
\providecommand{\BIBforeignlanguage}[2]{{%
\expandafter\ifx\csname l@#1\endcsname\relax
\typeout{** WARNING: IEEEtran.bst: No hyphenation pattern has been}%
\typeout{** loaded for the language `#1'. Using the pattern for}%
\typeout{** the default language instead.}%
\else
\language=\csname l@#1\endcsname
\fi
#2}}
\providecommand{\BIBdecl}{\relax}
\BIBdecl

\bibitem{b:CorkeVCbook97}
P.~I. Corke, \emph{Visual Control of Robots: High-Performance Visual
  Serving}.\hskip 1em plus 0.5em minus 0.4em\relax USA: John Wiley \& Sons,
  Inc., 1997.

\bibitem{b:Prasad2006FirstSI}
\BIBentryALTinterwordspacing
T.~D.~A. Prasad, K.~Hartmann, W.~Weihs, S.~E. Ghobadi, and A.~Sluiter, ``First
  steps in enhancing 3d vision technique using 2d/3d sensors,'' 2006. [Online].
  Available: \url{https://api.semanticscholar.org/CorpusID:17375912}
\BIBentrySTDinterwordspacing

\bibitem{b:bi2020automatic}
Z.~Bi, C.~Luo, Z.~Miao, B.~Zhang, and C.~W. Zhang, ``Automatic robotic
  recharging systems--development and challenges,'' \emph{Industrial Robot: the
  international journal of robotics research and application}, vol.~48, no.~1,
  pp. 95--109, 2020.

\bibitem{b:chen2022computer}
R.~Chen, C.~Zhou, and L.-l. Cheng, ``Computer-vision-guided semi-autonomous
  concrete crack repair for infrastructure maintenance using a robotic arm,''
  \emph{AI in Civil Engineering}, vol.~1, no.~1, p.~9, 2022.

\bibitem{b:BonchisTASE14}
A.~Bonchis, E.~Duff, J.~Roberts, and M.~Bosse, ``Robotic explosive charging in
  mining and construction applications,'' \emph{IEEE Transactions on Automation
  Science and Engineering}, vol.~11, no.~1, pp. 245--250, 2014.

\bibitem{b:SHIRAI197399}
\BIBentryALTinterwordspacing
Y.~Shirai and H.~Inoue, ``Guiding a robot by visual feedback in assembling
  tasks,'' \emph{Pattern Recognition}, vol.~5, no.~2, pp. 99--108, 1973.
  [Online]. Available:
  \url{https://www.sciencedirect.com/science/article/pii/0031320373900150}
\BIBentrySTDinterwordspacing

\bibitem{b:HaugaarDPML21}
\BIBentryALTinterwordspacing
R.~Haugaard, J.~Langaa, C.~Sloth, and A.~Buch, ``Fast robust peg-in-hole
  insertion with continuous visual servoing,'' in \emph{Proceedings of the 2020
  Conference on Robot Learning}, ser. Proceedings of Machine Learning Research,
  J.~Kober, F.~Ramos, and C.~Tomlin, Eds., vol. 155.\hskip 1em plus 0.5em minus
  0.4em\relax PMLR, 16--18 Nov 2021, pp. 1696--1705. [Online]. Available:
  \url{https://proceedings.mlr.press/v155/haugaard21a.html}
\BIBentrySTDinterwordspacing

\bibitem{b:TriyonoputroIROS19}
J.~C. Triyonoputro, W.~Wan, and K.~Harada, ``Quickly inserting pegs into
  uncertain holes using multi-view images and deep network trained on synthetic
  data,'' in \emph{2019 IEEE/RSJ International Conference on Intelligent Robots
  and Systems (IROS)}, 2019, pp. 5792--5799.

\bibitem{b:Chaumette06}
F.~Chaumette and S.~Hutchinson, ``Visual servo control. i. basic approaches,''
  \emph{IEEE Robotics and Automation Magazine}, vol.~13, no.~4, pp. 82--90,
  2006.

\bibitem{b:heikkila2010interactive}
T.~Heikkil{\"a} and J.~M. Ahola, ``An interactive 3d sensor system and its
  programming for target localizing in robotics applications author (s)
  heikkil{\"a}, ta,'' 2010.

\bibitem{b:XingTII23}
S.~Xing, F.~Jing, and M.~Tan, ``Reconstruction-based hand–eye calibration
  using arbitrary objects,'' \emph{IEEE Transactions on Industrial
  Informatics}, vol.~19, no.~5, pp. 6545--6555, 2023.

\bibitem{b:PetersJFR23}
\BIBentryALTinterwordspacing
A.~Peters and A.~C. Knoll, ``Robot self-calibration using actuated 3d
  sensors,'' \emph{Journal of Field Robotics}, vol.~41, no.~2, pp. 327--346,
  2024. [Online]. Available:
  \url{https://onlinelibrary.wiley.com/doi/abs/10.1002/rob.22259}
\BIBentrySTDinterwordspacing

\bibitem{b:MorrowICRA97}
J.~Morrow and P.~Khosla, ``Manipulation task primitives for composing robot
  skills,'' in \emph{Proceedings of International Conference on Robotics and
  Automation}, vol.~4, 1997, pp. 3354--3359 vol.4.

\bibitem{b:HutchinstonCorke96}
S.~Hutchinson, G.~Hager, and P.~Corke, ``A tutorial on visual servo control,''
  \emph{IEEE Transactions on Robotics and Automation}, vol.~12, no.~5, pp.
  651--670, 1996.

\bibitem{b:PrzystupaICRA21}
M.~Przystupa, M.~Dehghan, M.~Jagersand, and A.~R. Mahmood, ``Analyzing neural
  jacobian methods in applications of visual servoing and kinematic control,''
  in \emph{2021 IEEE International Conference on Robotics and Automation
  (ICRA)}, 2021, pp. 14\,276--14\,283.

\bibitem{b:HeikkilaSICE88}
T.~Heikkilä, T.~Matsushita, and T.~Sato, ``A line based object localizing
  method for robot vision,'' in \emph{SICE Robotics Meeting RS-88-9-11}, 1988,
  pp. 21--30.

\bibitem{b:HeikkilaSPIE99}
\BIBentryALTinterwordspacing
T.~A. Heikkila, M.~Sallinen, and M.~Jarviluoma, ``{Robot-based surface
  inspection system with estimation of spatial uncertainties},'' in
  \emph{Intelligent Robots and Computer Vision XVIII: Algorithms, Techniques,
  and Active Vision}, D.~P. Casasent, Ed., vol. 3837, International Society for
  Optics and Photonics.\hskip 1em plus 0.5em minus 0.4em\relax SPIE, 1999, pp.
  242 -- 253. [Online]. Available: \url{https://doi.org/10.1117/12.360304}
\BIBentrySTDinterwordspacing

\bibitem{b:Ahola2017}
J.~M. Ahola and T.~Heikkilä, ``Object recognition and pose estimation based on
  combined use of projection histograms and surface fitting,'' in
  \emph{Proceedings of the ASME Design Engineering Technical Conference},
  vol.~9, 2017.

\end{thebibliography}

\end{document}